\documentstyle[gb4e,11pt, twocolumn,lsalike]{article}

\newcommand{\ul}{\underline}
\newcommand{\bad}{\trans \makebox[10mm][l]{\sc old:}}
\newcommand{\good}{\vspace{-6mm} \trans \makebox[10mm][l]{\sc new:}}

\newcommand{\border}{\rule{\textwidth}{0.3mm}}
\newcommand{\halfborder}{\rule{\columnwidth}{0.3mm}}

\begin{document}

\title{Countability and Number\\ in Japanese-to-English Machine
  Translation}

\author{Francis {\sc Bond}, Kentaro {\sc Ogura}, Satoru {\sc Ikehara}\\
  {\bf NTT Communication Science Laboratories} \\
   1-2356 Take, Yokosuka-shi, Kanagawa-ken, {\sc Japan} 238-03 \\
    {\tt \{bond,ogura,ikehara\}@nttkb.ntt.jp}}

\date{COLING~94, August 1994\footnotemark[1]}


\maketitle

\begin{abstract}
  This paper presents a heuristic method that uses information in the
  Japanese text along with knowledge of English countability and
  number stored in transfer dictionaries to determine the countability
  and number of English noun phrases.  Incorporating this method into
  the machine translation system {\bf ALT-J/E}, helped to raise the
  percentage of noun phrases generated with correct use of articles
  and number from 65\% to 73\%.
\end{abstract}

\renewcommand{\thefootnote}{\fnsymbol{footnote}}
\footnotetext[1]{This paper was presented at COLING~94 and
  appears in the proceedings: Vol I, pp 32--38.}
\renewcommand{\thefootnote}{\arabic{footnote}}

\section{Introduction}
\label{sec:intro}

Correctly determining number is a difficult problem when translating
from Japanese to English.  This is because in Japanese, noun phrases
are not normally marked with respect to number.  Japanese nouns have
no equivalent to the English singular and plural forms and verbs do
not inflect to agree with the number of the subject \cite{Kuno:1973}.
In addition, there is no grammatical marking of
countability.\footnote{Japanese does not have obligatory plural
  morphemes.  Plurality can be marked but only rarely is, for example
by adding a suffix such as {\it tachi\/} ``and others'' (this can
normally only be used with people or animals).}

In order to generate English correctly, it is necessary to know
whether a given noun phrase is countable or uncountable and, if
countable, whether it is singular or plural.  Deciding this is a
problem even for humans translating from Japanese to English, but they
have their own knowledge of both languages to draw on.  A machine
translation system needs to have this knowledge codified in some way.
As generating articles and number is only important when the rest of
the sentence has been correctly generated, there has not been a lot of
research devoted to it.  Recently, \namecite{Murata:1993a} have
proposed a method of determining the referentiality property and
number of nouns in Japanese sentences for machine translation into
English, but the research has not yet been extended to include the
actual English generation.

This paper describes a method that extracts information relevant to
countability and number from the Japanese text and combines it with
knowledge about countability and number in English.  First countability
in English is discussed at the noun phrase and then the noun level.
As a noun phrase's countability in English is affected by its
referential property (generic, referential or ascriptive) we present
a method of determining the referential use of Japanese noun phrases.
Next the process of actually determining noun phrase countability and
number is described.  This is followed by some examples of sentences
translated by the proposed method and a discussion of the results.

The processing described in this paper has been implemented in NTT
Communication Science Laboratories' experimental machine translation
system {\bf ALT-J/E} \cite{Ikehara:1991}.  Along with new processing
for the generation of articles, which is not discussed in detail in
this paper, it improved the percentage of noun phrases with correctly
generated determiners and number from 65\% to 73\%.

\section{Countability}
\label{sec:count}



\subsection{Noun Phrase Countability}

\begin{table*}[tp]
  \border
  \caption{Lexical information for nouns}
  \label{nounexamples}
  \begin{center}
  \begin{tabular}{|ll|lll|}
    \multicolumn{5}{c}{} \\ \hline
    \multicolumn{2}{|c|}{Noun}  & Countability  & Default  & Default \\
    English & Japanese  & Preference    & Number  &  Classifier \\ \hline
    knife & \it houchou & \sc fully countable & \sc singular & --- \\
    noodles & \it men & \sc fully countable & \sc plural & --- \\
    group & \it mure &  \sc (collective) & \sc singular  & --- \\
    cake & \it ke--ki &\sc strongly countable & \sc singular & --- \\
    beer & \it bi--ru &\sc weakly countable & \sc singular & --- \\
    furniture & \it kagu & \sc uncountable & \sc singular & \it piece \\
    knowledge & \it chishiki & \sc (semi-countable) & \sc singular & \it
piece\\
    scissors & \it hasami & \sc pluralia tantum & \sc plural & \it pair \\
    clothes & \it ifuku & \sc pluralia tantum & \sc plural & --- \\
    \hline
  \end{tabular}
  \end{center}
  \border
\end{table*}

We adopt the definition of countability in English given in
\namecite[541--3]{Allan:1980}.  A countable noun phrase is defined as
follows:
\begin{description}
\item[I] If the head constituent of an NP falls within the scope of a
  denumerator it is countable.
\item[II] If the head constituent of an NP is plural it is countable.
\end{description}

Where ``the phrase `falls within the scope [or domain] of a
denumerator' means `is denumerated' by it; i.e the NP reference is
quantified by the denumerator as a number of discrete entities.''

Not all nouns in English can become the head of a countable noun
phrase.  In particular, noun phrases whose heads fall within the scope
of a denumerator (`denumerated' noun phrases) must be headed by a noun
that has both singular and plural forms.  Nouns that do not have both
forms, like {\it equipment} or {\it scissors}, require a classifier to be
used.  The classifier becomes the head of a countable noun phrase with
the original noun attached as the complement of a prepositional
phrase headed by {\it of}: {\it a pair of scissors, a piece of equipment.}

Whether a noun can be used to head a countable noun phrase or not
depends both on how it is interpreted, and on its inherent
countability preference.  Noun countability preferences are discussed
in the next section.

\subsection{Noun Countability Preferences}

A noun's countability preference determines how it will behave in
different environments.  We classify nouns into seven countability
preferences, five major and two minor, as described below.

The two most basic types are `fully countable' and `uncountable'.
Fully countable nouns, such as {\it knife\/} have both singular and
plural forms, and cannot be used with determiners such as {\it
  much}.\footnote{The determiners {\it much, little, a little, less
    and overmuch.\/} can all be used for this test} Uncountable nouns,
such as {\it furniture}, have no plural form, and can be used with
{\it much.}

Between these two extremes there are a vast number of nouns, such as
{\it cake}, that can be used in both countable and uncountable noun
phrases.  They have both singular and plural forms, and can also be
used with {\it much.} Whether such nouns will be used countably or
uncountably depends on whether their referent is being thought of as
made up of discrete units or not.  As it is not always possible to
explicitly determine this when translating from Japanese to English,
we divide these nouns into two groups: `strongly countable', those
that are more often used to refer to discrete entities, such as {\it
  cake}, and `weakly countable', those that are more often used to
refer to unbounded referents, such as {\it beer}.

The last major type of countability preference is `pluralia tanta':
nouns that only have a plural form, such as {\it scissors}.  They can
neither be denumerated nor modified by {\it much.} We further
subdivide pluralia tanta into two types, those that can use the
classifier {\it pair\/} to be denumerated, such as {\it a pair of
  scissors\/} and those that can't, such as {\it clothes}.  `pair'
pluralia tanta have a singular form when used as modifiers ({\it a
  scissor movement}).  Pluralia tanta such as {\it clothes}, use the
plural form even as modifiers ({\it a clothes horse}), and need a
countable word of similar meaning to be substituted when they are
denumerated: {\it a garment, a suit, \ldots\/}.

The two minor types are subsets of fully countable and uncountable
nouns respectively.  Unless explicitly indicated, they will be treated
the same as their supersets.  `Collective' nouns share all the
properties of fully countable nouns.  In addition they can have
singular or plural verb agreement with the singular form of the noun:
{\it The government has/have decided}.  `Semi-countable' nouns share
the properties of uncountable nouns, except that they can be modified
directly by {\it a/an}; for example {\it a knowledge [of Japanese]}.

Examples of the information about countability and number stored in
the Japanese to English noun transfer dictionary are given in table
\ref{nounexamples}.  The information about noun countability
preferences cannot be found in standard dictionaries and must be
entered by an English native speaker.  Some tests to help determine a
given noun's countability preferences are described in
\namecite{Bond:1993a}, which discusses the use of noun countability
preferences in Japanese to English machine translation.

\section{Determination of NP Referentiality}
\label{sec:ref/gen}

The first stage in generating the countability and number of a
translated English noun phrase is to determine its referentiality.  We
distinguish three kinds of referentiality: `generic', `referential'
and `ascriptive'.

We call noun phrases used to make general statements about a class
generic; for example {\it \ul{Mammoths} are extinct}.  The way generic
noun phrases are expressed in English is described in Section
\ref{subsec:gen}.  Referential noun phrases are ones that refer to
some specific referent; for example {\it \ul{Two dogs} chase \ul{a
    cat}.} Their number and countability are ideally determined by the
properties of the referent.  Ascriptive noun phrases are used to
ascribe a property to something; for example {\it Hathi is \ul{an
    elephant}.} They normally have the same number and countability as
the noun phrase whose property they are describing.

\begin{figure}[htpb]
\halfborder
\small
\begin{enumerate}
\item \label{det:a}
  if restrictively modified then `referential'\\
  {\it \ul{my book}, \ul{the man who came to dinner}\/}
\item \label{det:b}
  if subject of {\it extinct, evolve\/} \ldots `generic'\\
  {\it \ul{Mammoths} are extinct}
\item \label{det:z}
  if the semantic category of the subject of a copula  is a daughter
  of the semantic category of the object then `generic'\\
  {\it \ul{Mammoths} are animals}
\item \label{det:c}
  if modified by {\it aimed at, for}\/ \ldots then `generic'\\
  {\it A magazine for \ul{women}\/}
\item \label{det:d}
  if object of {\it like}\/ \ldots then `generic'\\
  {\it I like \ul{cake}\/}
\item \label{det:e}
  if complement of a copula then `ascriptive'\\
  {\it NTT is \ul{a telephone company}}
\item \label{det:f}
  if appositive then `ascriptive'\\
  {\it NTT, \ul{a telephone company}} \ldots
\item \label{det:g}
  default `referential'
\end{enumerate}
\caption{Determination of NP referentiality}
\label{fig:ref}
\halfborder
\end{figure}

The process of determining the referentiality of a noun phrase is
shown in Figure \ref{fig:ref}.  The tests are processed in the order
shown.  As far as possible, simple criteria that can be implemented
using the dictionary have been chosen.  For example, Test \ref{det:c}
`` if a NP is modified by {\it aimed at, for}\/ \ldots then it is
`generic'\/'' is applied as part of translating NP1-{\it muke} into
``for NP1''. The transfer dictionary includes the information that in
this case, NP1 should be generic.

Tests \ref{det:b} a\ref{det:z} show two more heuristic methods for
determining whether a noun phrase has generic reference.  In Test
\ref{det:b}, if the predicate is marked in the dictionary as one that
only applies to classes as a whole, such as {\it evolve\/} or {\it be
  extinct\/}, then the sentence is taken to be generic.  In Test
\ref{det:z}, {\bf ALT-J/E}'s semantic hierarchy is used to test
whether a sentence is generic or not.  For example in {\it Mammoths
  are animals}, {\it mammoth\/} has the semantic category {\sc
  animal\/} so the sentence is judged to be stating a fact true of all
mammoths and is thus generic.

\subsection{Generic noun phrases}
\label{subsec:gen}

A generic noun phrase (with a countable head noun) can generally be
expressed in three ways \cite{Huddleston:1984}.  We call these GEN
`a', where the noun phrase is indefinite: {\it A mammoth is a mammal};
GEN `the', where the noun phrase is definite: {\it The mammoth is a
  mammal}; and GEN $\phi$, where there is no article: {\it Mammoths
  are mammals}.  Uncountable nouns and pluralia tanta can only be
expressed by GEN $\phi$ (eg: {\it Furniture is expensive}).  They
cannot take GEN `a' because they cannot be modified by {\it a}.  They
do not take GEN `the', because then the noun phrase would normally be
interpreted as having definite reference.  Nouns that can be either
countable or uncountable also only take GEN $\phi$: {\it Cake is
  delicious}/{\it Cakes are delicious}.  These combinations are shown
in Table \ref{gandc}, noun phrases that can not be used to show
generic reference are marked *.

\begin{table}[htp]
\halfborder
\caption{Genericness and Countability} \label{gandc}
\begin{center}
\footnotesize \begin{tabular}{|c|l|l|l|} \hline
GEN     & \multicolumn{3}{|c|}{Noun Countability Preference}  \\
\cline{2-4}
type    & Countable     & Both  & Uncountable    \\ \hline
`a'     & a mammoth     & *a cake       & *a furniture  \\
`the'   & the mammoth   & *the cake     & *the furniture  \\
$\phi$  & mammoths      & cake/cakes    & furniture      \\
\hline
\end{tabular}
\end{center}
\halfborder
\end{table}

The use all three kinds of generic noun phrases is not acceptable in
some contexts, for example * {\it a mammoth evolved}.  Sometimes a
noun phrase can be ambiguous, for example {\it I like the elephant},
where the speaker could like a particular elephant, or all elephants.

Because the use of GEN $\phi$ is acceptable in all contexts, {\bf
  ALT-J/E} generates all generic noun phrases as such, that is as bare
noun phrases.  The number of the noun phrase is then determined by the
countability preference of the noun phrase heading it.  Fully
countable nouns and pluralia tanta will be plural, all others are
singular.

\section{Determination of NP Countability and Number}
\label{sec:M/I}

The following discussion deals only with referential and ascriptive noun
phrases as generic noun phrases were discussed in Section~\ref{subsec:gen},

\begin{figure}[htbp]
\halfborder  \small
  \begin{enumerate}
  \item if the Japanese is explicitly plural then \\
    countable and plural
  \item determine according to determiner \\
    {\it one dog, all dogs\/}
  \item determine according to classifier \\
    {\it a slice of cake, a pile of cakes\/}
  \item determine according to complement \\
    {\it schools all over the country\/}
  \item ascriptive NPs match their subjects \\
    {\it A computer is a piece of equipment\/}
  \item determine according to verb \\
    {\it I gather flowers\/}
  \item use default value
    \begin{enumerate}
    \item uncountable, weakly countable become: \\
      uncountable and singular
    \item pluralia tanta become: \\
      countable and plural
      \item countable and strongly countable become: \\
        countable and singular or plural \\
        according to the dictionary default
    \end{enumerate}
  \end{enumerate}
  \caption{Determination of English noun phrase Countability and Number}
  \label{fig:count}
\halfborder
\end{figure}

\begin{table*}[tp]
\border
\caption{Noun Phrase Countability and Number}
\label{tab:count}
\begin{center}

\small
\begin{tabular}{|l|l|l|l|l|}
\hline
Noun Type &  \multicolumn{2}{|c|}{Denumerated} &
\multicolumn{2}{|c|}{Mass} \\  \cline{2-5}
 & Singular & Plural & Countable & Uncountable \\ \hline
 Fully Countable  &  a dog  & two dogs & dogs & dogs \\
 Strongly Countable   &  a cake & two cakes & cakes & cake \\
 Weakly Countable &  a beer & two beers & beer & beer \\
 Uncountable      &  a piece of information  & two pieces of
 information & information & information \\
 Pluralia Tantum  &  a pair of scissors &  two pairs of scissors &
 scissors & scissors \\
\hline
\end{tabular}
\end{center}
\border
\end{table*}

The definitions of noun phrase countability given in Section
\ref{sec:count}, while useful for analyzing English, are not
sufficient for translating from Japanese to English.  This is because
in many cases it is impossible to tell from the Japanese form or
syntactic shape whether a translated noun phrase will fall within the
scope of a denumerator or not.  Japanese has no equivalent to {\it
  a/an} and does not distinguish between countable and uncountable
quantifiers such as {\it many/much\/} and {\it little/few}.  Therefore
to determine countability and generate number we need to use a
combination of information from the Japanese original sentence, and
default information from the Japanese to English transfer dictionary.
As much as possible, detailed information is entered in the transfer
dictionaries to allow the translation process itself to be made
simple.

The process of determining a noun phrase's countability and number is
shown in Figure \ref{fig:count}.  The process is carried out during
the transfer stage so information is available from both the Japanese
original and the selected English translation.

To make the task of determining countability and number simpler, we
define combinations of different countabilities for nouns with
different countability preferences that we can use in the
dictionaries.  The effects of the four most common types on the five
major noun countability preferences are shown in Table \ref{tab:count}.

Noun phrases modified by Japanese/English pairs that are translated as
denumerators we call denumerated.  For example a noun modified by {\it
  onoono-no\/} ``each'' is denumerated - singular, while one modified
by {\it ryouhou-no\/} ``both'' is denumerated - plural.  Uncountable
and pluralia tantum nouns in denumerated environments are translated
as the prepositional complement of a classifier.  A default classifier
is stored stored in the dictionary for uncountable nouns and pluralia
tanta.  Ascriptive noun phrases whose subject is countable will also
be denumerated.

The two `mass'\footnote{We called these environments `mass' because
  they both can be used to show a mass or unbounded interpretation.}
environments shown in Table~\ref{tab:count} are used to show the
countability of nouns that can be either countable or uncountable.
Weakly countable nouns will only be countable if used with a
denumerator.  Strongly countable nouns will be countable and plural in
such mass - countable environments as the object of {\it collect\/}
(vt): {\it I collect \ul{cakes}}, and uncountable and singular in mass
-uncountable environments such as {\it I ate too much cake\/}.  In
fact, both {\it I collect cake\/} and {\it I ate too many cakes\/} are
possible.  As Japanese does not distinguish between the two the system
must make the best choice it can, in the same way a human translator
would have to.  The rules have been implemented to generate the
translation that has the widest application, for example generating
{\it I ate too much cake}, which is true whether the speaker only ate
part or all of one cake or if they ate many cakes, rather than {\it I
  ate too many cakes\/} which is only true if the speaker ate many
cakes.

Sometimes the choice of the English translation of a modifier will
depend on the countability of the noun phrase.  For example, {\it
  kazukazu-no} and {\it takusan-no\/} can all be translated as
``many''.  {\it kazukazu-no\/} implies that it's modificant is made up
of discrete entities, so the noun phrase it modifies should be
translated as denumerated - plural.  {\it takusan-no\/} does not carry
this nuance so {\bf ALT-J/E} will translate a noun phrase modified by
it as mass - uncountable, and {\it takusan-no\/} as {\it many\/} if
the head is countable and {\it much\/} otherwise.

Rules that translate the nouns with different noun countability
preferences into other combinations of countable and uncountable are
also possible.  For example, sometimes even fully countable nouns can
be used in uncountable noun phrases.  If an elephant is referred to
not as an individual elephant but as a source of meat, then it will be
expressed in an uncountable noun phrase: {\it I ate a slice of
  elephant}.  To generate this the following rule is used: ``nouns
quantified with the classifier {\it kire\/} ``slice'' will be
generated as the prepositional complement of {\it slice\/}, they will
be singular with no article unless they are pluralia tanta, when they
will be plural with no article''.

Note that countable indefinite singular noun phrases without a
determiner will have {\it a/an\/} generated.  Countable indefinite
plural noun phrases and uncountable noun phrases may have {\it some\/}
generated; a full discussion of this is outside the scope of this
article.

\section{Experimental Results}
\label{sec:results}

This processing described above has been implemented in {\bf ALT-J/E}.
It was tested, together with new processing to generate articles, on a
specially constructed set of test sentences, and on a collection of
newspaper articles.  The results are summarized in Table
\ref{tab:results}.

\begin{table}[htbp]
  \halfborder
  \caption{Correct Generation of Articles and Number}
  \label{tab:results}
  \begin{center}
    \footnotesize \begin{tabular}{|l|c|c|c|c|} \hline
      & \multicolumn{2}{|c|}{Test Sentences}
      & \multicolumn{2}{|c|}{Newspaper Articles} \\ \cline{2-5}
      & NPs   & Sentences     & NPs   & Sentences \\
      & (240) & (120)         & (717) & (102)  \\ \hline
      New    & 94\%  & 90\%          & 73\%  & 12\%  \\
      Old    & 70\%  & 46\%          & 65\%  & 5\%  \\ \hline
    \end{tabular}
  \end{center}
  \halfborder
\end{table}

In the newspaper articles tested, there were an average of 7.0 noun
phrases in each sentence.  For a sentence to be judged as correct all
the noun phrases must be correct.  The introduction of the proposed
method improved the percentage of correct sentences from 5\% to 12\%.

Some examples of translations before and after the introduction of the
new processing are given below.  The translations before the proposed
processing was implemented are marked {\sc old}, the translations
produced by {\bf ALT-J/E} using the proposed processing are marked
{\sc new}.

\begin{exe}
\ex \label{ex:asc}
\gll   taitei-no kodomo-ha \ul{otona}-ni naru \\
       most child \ul{adult} become \\
\bad   ``Most children become \ul{an adult}'' \\
\good   ``Most children become \ul{adults}''
\end{exe}

In (\ref{ex:asc}), the noun phrase headed by {\it otona\/} ``adult'' is
judged to be prescriptive, as it is the complement of the copular {\it
  naru\/} ``become''.  Therefore the proposed method translates it
with the same number as the subject.

\begin{exe}
\ex \label{ex:gen}
\gll   \ul{manmosu}-ha zetumetu-shita \\
       \ul{mammoth} died-out \\
\bad   ``\ul{A mammoth} died out'' \\
\good  ``\ul{Mammoths} died out''
\end{exe}

{\it zetumetu\/} ``die out'', is entered in the lexicon as a verb
whose subject must be generic.  {\it manmosu\/} ``mammoth'' is fully
countable so the generic noun phrase is translated as a bare plural.

\begin{exe}
\ex
\gll   \ul{tofu 3-chou}, \ul{hasami 1-chou}, \ul{houchou 2-chou-ga}
       aru \\
       \ul{tofu 3}, \ul{scissors 1}, \ul{knife 2} is \\
\bad   ``There are \ul{3 piece tofu}, \ul{1 scissors},
       and \ul{2 knives}'' \\
\good  ``There are \ul{3 pieces of tofu}, \ul{1 pair}
       \ul{of scissors} and \ul{2 knives}''
\end{exe}

The old version recognizes that a denumerated noun phrase headed by an
uncountable noun {\it tofu\/} requires a classifier but does not
generate the correct structure neither does it generate a classifier
for the pluralia tanta {\it scissors\/}.  The version using the
proposed method does.

\begin{exe}
\ex
\gll   sore-ha \ul{dougu} da \\
       that \ul{equipment} is \\
\bad   ``That is \ul{equipment}'' \\
\good  ``That is  \ul{a piece of equipment}''
\end{exe}

As the subject of the copula {\it that\/} is countable it's complement
is judged to be denumerated by the proposed method.  As the complement
is headed by an uncountable noun it must be embedded in the
prepositional complement of a classifier.

There are three main problems still remaining.  The first is that
currently the rules for determining the noun phrase referentiality are
insufficiently fine.  We estimate that if referentiality could be
determined 100\% correctly then the percentage of noun phrases with
correctly generated articles and number could be improved to 96\% in
the test set we studied.  The remaining 4\% require knowledge from
outside the sentence being translated.  The biggest problem is noun
phrases requiring world knowledge that cannot be expressed as a
dictionary default.  These noun phrases cannot be generated correctly
by the purely heuristic methods proposed here.  The last problem is
noun phrases whose countability and number can be deduced from
information in other sentences.  We would like to extend our method to
use this information in the future.


\section{Conclusion}
\label{sec:conc}

The quality of the English in a
Japanese to English Machine Translation system can be improved by the
method proposed in this paper.  This method uses the information
available in the original Japanese sentence along with information
about English countability at both the noun phrase and noun level that
can be stored in Japanese to English transfer dictionaries.
Incorporating this method into the machine translation system
 \mbox{{\bf ALT-J/E}} helped to improve the percentage of noun phrases
with correctly generated articles and number from 65\% to 73\%.


\begin{thebibliography}{}

\bibitem[\protect\citeauthoryear{Allan}{1980}]{Allan:1980}
{\sc Allan, Keith}.
\newblock 1980.
\newblock Nouns and countability.
\newblock {\em Language\/} 56.541--67.

\bibitem[\protect\citeauthoryear{Bond \lsaand Ogura}{1993}]{Bond:1993a}
{\sc Bond, Francis},  \lsaand {\sc Kentaro Ogura}.
\newblock 1993.
\newblock Determination of whether an {English} noun phrase is countable or not
  using 6 levels of lexical countability.
\newblock In {\em Proceedings of the 46th Annual Convention IPSJ Japan\/},
  6:107--108.
\newblock (in {Japanese}).

\bibitem[\protect\citeauthoryear{Huddleston}{1984}]{Huddleston:1984}
{\sc Huddleston, Rodney}.
\newblock 1984.
\newblock {\em Introduction to the Grammar of English\/}.
\newblock Cambridge textbooks in linguistics. Cambridge: Cambridge University
  Press.

\bibitem[\protect\citeauthoryear{Ikehara {\em et~al.\/}}{1991}]{Ikehara:1991}
{\sc Ikehara, Satoru}, {\sc Satoshi Shirai}, {\sc Akio Yokoo},  \lsaand {\sc
  Hiromi Nakaiwa}.
\newblock 1991.
\newblock Toward an {MT} system without pre-editing -- effects of new methods
  in {{\bf ALT-J/E}}--.
\newblock In {\em Proceedings of MT Summit III\/}, 101--106.
\newblock (cmp-lg/9510008).

\bibitem[\protect\citeauthoryear{Kuno}{1973}]{Kuno:1973}
{\sc Kuno, Susumu}.
\newblock 1973.
\newblock {\em The Structure of the {Japanese} Language\/}.
\newblock Cambridge, Massachusetts, and London, England: MIT Press.

\bibitem[\protect\citeauthoryear{Murata \lsaand Nagao}{1993}]{Murata:1993a}
{\sc Murata, Masaki},  \lsaand {\sc Makoto Nagao}.
\newblock 1993.
\newblock Determination of referential property and number of nouns in
  {Japanese} sentences for machine translation into {English}.
\newblock In {\em Proceedings of the Fifth International Conference on
  Theoretical and Methodological Issues in Machine Translation (TMI-93)\/},
  218--25.

\end{thebibliography}

\end{document}